# A Framework for Decision-Theoretic Planning I: Combining the Situation Calculus, Conditional Plans, Probability and Utility*

David Poole
Department of Computer Science
University of British Columbia
Vancouver, B.C., Canada V6T 1Z4
poole@cs.ubc.ca
http://www.cs.ubc.ca/spider/poole

## Abstract

This paper shows how we can combine logical representations of actions and decision theory in such a manner that seems natural for both. In particular we assume an axiomatization of the domain in terms of situation calculus, using what is essentially Reiter's solution to the frame problem, in terms of the completion of the axioms defining the state change. Uncertainty is handled in terms of the independent choice logic, which allows for independent choices and a logic program that gives the consequences of the choices. As part of the consequences are a specification of the utility of (final) states. The robot adopts robot plans, similar to the GOLOG programming language. Within this logic, we can define the expected utility of a conditional plan, based on the axiomatization of the actions, the uncertainty and the utility. The 'planning' problem is to find the plan with the highest expected utility. This is related to recent structured representations for POMDPs; here we use stochastic situation calculus rules to specify the state transition function and the reward/value function. Finally we show that with stochastic frame axioms, actions representations in probabilistic STRIPS are exponentially larger than using the representation proposed here.

## 1 Introduction

The combination of decision theory and planning is very appealing. Since the combination was advocated in [Feldman and Sproull, 1975], there has been a recent revival of interest. The general idea of planning is to construct a sequence of steps, perhaps conditional on observations that solves a goal. In decision-theoretic planning, this is generalised to the case where there is uncertainty about the environment and we are concerned, not only with solving a 'goal', but what happens under any of the contingencies. Goal solving is extended to the problem of maximizing the agent's expected utility, where here the utility is an arbitrary function of the final state.

Recently there have been claims made that Markov decision processes (MDPs) [Puterman, 1990] are the appropriate framework for developing decision theoretic planners (e.g., [Boutilier and Puterman, 1995; Dean et al., 1993]). In MDPs the idea is to construct policies — functions from observed state into actions that maximise long run cumulative (usually discounted) reward. It is wrong to equate decision theory with creating policies; decision theory can be used to select plans, and policies can be considered independently of decision theory [Schoppers, 1987]. Even when solving partially observable MDPs (POMDPs), where a policy is a function from belief states into actions, it is often more convenient to use a policy tree [Kaelbling et al., 1996], which is much more like a robot plan as developed here — see Section 7.

Rather than assuming robots have policies [Poole, 1995c], we can instead consider robot plans as in GOLOG [Levesque et al., 1996]. These plans consider sequences of steps, with conditions and loops, rather than reactive strategies. In this paper we restrict ourselves to conditional plans; we do not consider loops or nondeterministic choice, although these also could be considered (see Section 6). Unlike GOLOG, and like Levesque [1996], the conditions in the branching can be 'observations' about the world or values received by sensors

As in GOLOG, we assume that the effects of actions are represented in the situation calculus. In particular we adopt Reiter's [1991] solution to the frame problem. Our representation is simpler in that we do not assume that actions have preconditions — all actions can be attempted at any time, the effects of these actions can depend on what else is true in the world. This is important because the agent may not know whether the preconditions of an action hold, but, for example, may be sure enough to want to try the action.

*This work was supported by Institute for Robotics and Intelligent Systems, Project IC-7 and Natural Sciences and Engineering Research Council of Canada Operating Grant OGP0044121. Thanks to Craig Boutilier, Ronen Brafman and Chris Geib for giving me a hard time about this paper.

All of the uncertainty in our rules is relegated to independent choices as in the independent choice logic [Poole, 1995b] (an extension of probabilistic Horn abduction [Poole, 1993]). This allows for a clean separation of the completeness assumed by Reiter's solution to the frame problem and the uncertainty we need for decision theory.

Before we describe the theory there are some design choices incorporated into the framework:

- In the deterministic case, the trajectory of actions by the (single) agent up to some time point determines what is true at that point. Thus, the trajectory of actions, as encapsulated by the 'situation' term of the situation calculus [McCarthy and Hayes, 1969; Reiter, 1991] can be used to denote the state, as is done in the traditional situation calculus. However, when dealing with uncertainty, the trajectory of an agent's actions up to a point, does not uniquely determine what is true at that point. What random occurrences or exogenous events occurred also determines what is true. We have a choice: we can keep the semantic conception of a situation (as a state) and make the syntactic characterization more complicated by perhaps interleaving exogenous actions, or we can keep the simple syntactic form of the situation calculus, and use a different notion that prescribes truth values. We have chosen the latter, and distinguish the 'situation' denoted by the trajectory of actions, from the 'state' that specifies what is true in the situation. In general there will be a probability distribution over states resulting from a set of actions by the agent. It is this distribution over states, and their corresponding utility, that we seek to model.

    This division means that agent's actions are treated very differently from exogenous actions that can also change what is true. The situation terms define only the agent's actions in reaching that point in time. The situation calculus terms indicate only the trajectory, in terms of steps, of the agent and essentially just serve to delimit time points at which we want to be able to say what holds.

- When building conditional plans, we have to consider what we can condition these plans on. We assume that the agent has passive sensors, and that it can condition its actions on the output of these sensors. We only have one sort of action, and these actions only affect 'the world' (which includes both the robot and the environment). All we need to do is to specify how the agent's sensors depend on the world. This does not mean that we cannot model information-producing actions (e.g., looking in a particular place) — these information producing actions produce effects that make the sensor values correlate with what is true in the world. The sensors can be noisy — the value they return does not necessarily correspond with what is true in the world (of course if there was no correlation with what is true in the world, they would not be very useful sensors).

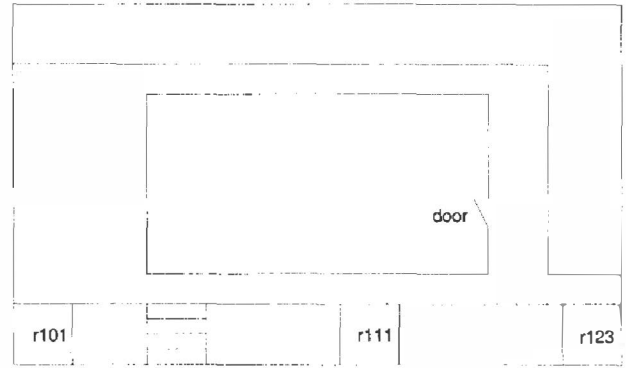

Figure 1: The example robot environment

- When mixing logic and probability, one can extend a rich logic with probability, and have two sorts of uncertainty — that uncertainty from the probabilities and that from disjunction in the logic [Bacchus, 1990]. An alternative that is pursued in the independent choice logic is to have all of the uncertainty in terms of probabilities. The logic is restricted so that there is no uncertainty in the logic — every set of sentences has a unique model. In particular we choose the logic of acyclic logic programs under the stable model semantics; this seems to be the strongest practical language with the unique model property. All uncertainty is handled by what can be seen as independent stochastic mechanisms, and a deterministic logic program that gives the consequences of the agent's actions and the random outcomes. In this manner we can get a simple mix of logic and Bayesian decision theory (see [Poole, 1995b]).

- Unlike in Markov decision processes, where there is a reward for each state and utilities accumulate, we assume that an agent carries out a plan, and receives utility depending on the state it ends up in. This is done to simplify the formalism, and seems natural. This does not mean that we cannot model cases where an agent receives rewards and costs along the way, but the rewards accumulated then have to be part of the state. Note that MDPs also need to make the cumulative reward part of the state to model non-additive rewards such as an agent receiving or paying interest on its current utility. This also means that we cannot optimize ongoing processes that never halt — in fact the acyclic restriction in the language means that we cannot model such ongoing processes without inventing an arbitrary stopping criteria (e.g., stop after 3 years).

We use the following ongoing example to show the power of the formalism; it is not intended to be realistic.

**Example 1.1** Suppose we have a robot that can travel around an office building, pick up keys, unlock doors, and sense whether the key is at the location it is currently at. In the domain depicted in Figure 1, we assume we want to enter the lab, and there is uncertainty about whether the door is locked or not, and uncertainty about where the key is (and moreover the probabilities are not independent). There are also stairs that the robot can fall down, but it can choose to go around the long way rather and avoid the stairs. The utility of a plan depends on whether it gets into the lab, whether it falls down the stairs and the resources used.

## 2 The Situation Calculus and a Solution to the Frame Problem

Before we introduce the probabilistic framework we present the situation calculus [McCarthy and Hayes, 1969] and a simple solution to the frame problem due to Kowalski [Kowalski, 1979], Schubert [Schubert, 1990] and Reiter [Reiter, 1991].

The general idea is that robot actions take the world from one 'situation' to another situation. We assume there is a situation $s_0$ that is the initial situation, and a function $do(A, S)$[1] that given action $A$ and a situation $S$ returns the resulting situation.

**Example 2.1** $do(goto(room111), s_0)$ is a situation resulting from the agent attempting to go to room 111 from situation $s_0$. $do(pickup(key), do(goto(room111), s_0))$ is the situation resulting from the agent attempting to pick up a key after it has attempted to go to room 111.

For this paper a situation is defined in terms of the $s_0$ constant and the $do$ function. An agent that knows what it has done, knows what situation it is in. It however does not necessarily know what is true in that situation. The robot may be uncertain about what is true in the initial situation, what the effects of its actions are and what exogenous events occurred.

We model all randomness as independent stochastic mechanisms, such that an external viewer that knew the initial state (i.e., what is true in the situation $s_0$), and knew how the stochastic mechanisms resolved themselves would be able to predict what was true in any situation. Given a probability distribution over the stochastic mechanisms, we have a probability distribution over the effects of actions.

We will use logic to specify the transitions specified by actions and thus what is true in a situation. What is true in a situation depends on the action attempted, what was true before and what stochastic mechanism occurred. A fluent is a predicate (or function) whose value depends on the situation; we will use the situation as the last argument to the predicate (function). We assume that for each fluent we can axiomatise in what situations it is true based on the action that was performed, what was true in the previous state and the outcome of the stochastic mechanism.

[1] We assume the Prolog convention that variables are in upper case and constants are in lower case. Free variables in formulae are considered to be universally quantified with the widest scope.

**Example 2.2** We can write rules such as, the robot is carrying the key after it has (successfully) picked it up:

$$carrying(key, do(pickup(key), S)) \leftarrow \\ at(robot, Pos, S) \land \\ at(key, Pos, S) \land \\ pickup\_succeeds(S).$$

Here $pickup\_succeeds(S)$ is true if the agent would succeed if it picks up the key and is false if the agent would fail to pick up the key. The agent typically does not know the value of $pickup\_succeeds(S)$ in situation $S$, or the position of the key.

The general form of a frame axiom specifies that a fluent is true after a situation if it were true before, and the action were not one that undid the fluent, and there was no mechanism that undid the fluent.

**Example 2.3** The agent is carrying the key as long as the action was not to put down the key or pick up the key, and the agent did not accidentally drop the key while carrying out another action[2]:

$$carrying(key, do(A, S)) \leftarrow \\ carrying(key, S) \land \\ A \neq putdown(key) \land \\ A \neq pickup(key) \land \\ keeps\_carrying(key, S).$$

Like $pickup\_succeeds(S)$ in Example 2.2, $keeps\_carrying(key, S)$ may be something that the agent does not know whether it is true — there may be a probability that the agent will drop the key. This thus forms a stochastic frame axiom. Note that the same mechanism that selects between dropping the key and keeping on carrying the key may also have other effects.

We assume that the clauses are acyclic [Apt and Bezem, 1991]. Recursion is allowed but all recursion much be well founded. The clauses represent the complete description of when the predicate will hold.

## 3 The Independent Choice Logic

The Independent Choice Logic specifies a way to build possible worlds. Possible worlds are built by choosing propositions from independent alternatives, and then extending these 'total choices' with a logic program. This section defines the logic $ICL_{SC}$.

[2] Note that $A \neq pickup(key)$ is a condition here to cover the case where the robot is holding the key and attempts to pick it up. With the inequality the robot has the same chance of succeeding as a pickup action when the agent is not holding the key. Without this condition, the agent would not be holding the key only if it dropped the key and the pickup failed.

Note that a possible world correspond to a complete history. A possible world will specify what is true in each situation. In other words, given a possible world and a situation, we can determine what is true in that situation. We define the independent choice logic without reference to situations — the logic programs will refer to situations.

There are two languages we will use: $\mathcal{L}_F$ which, for this paper, is the language of acyclic logic programs [Apt and Bezem, 1991], and the language $\mathcal{L}_Q$ of queries which we take to be arbitrary propositional formulae (the atoms corresponding to ground atomic formulae of the language $\mathcal{L}_F$). We write $f \hspace{1pt}\vert\hspace{-4pt}\sim q$ where $f \in \mathcal{L}_F$ and $q \in \mathcal{L}_Q$ if $q$ is true in the unique stable model of $f$ or, equivalently, if $q$ follows from Clark's completion of $q$ (the uniqueness of the stable model and the equivalence for acyclic programs are proved in [Apt and Bezem, 1991]). See [Poole, 1995a] for a detailed analysis of negation as failure in this framework, and for an abductive characterisation of the logic.

**Definition 3.1** A **choice space** is a set of sets of ground atomic formulae, such that if $C_1$, and $C_2$ are in the choice space, and $C_1 \neq C_2$ then $C_1 \cap C_2 = \{\}$. An element of a choice space is called a **choice alternative** (or sometimes just an alternative). An element of a choice alternative is called an **atomic choice**. An atomic choice can appear in at most one alternative.[3]

**Definition 3.2** An ICL$_{SC}$ **theory** is a tuple $\langle \mathcal{C}_0, A, \mathcal{O}, P_0, \mathcal{F} \rangle$ where

- $\mathcal{C}_0$ called **nature's choice space**, is the choice space of alternatives controlled by nature.

- $A$ called the **action space**, is a set of primitive actions that the agent can perform.

- $\mathcal{O}$ the **observables** is a set of terms.

- $P_0$ is a function $\cup \mathcal{C}_0 \to [0,1]$ such that $\forall C \in \mathcal{C}_0$, $\sum_{c \in C} P_0(c) = 1$. I.e., $P_0$ is a probability measure over the alternatives controlled by nature.

- $\mathcal{F}$ called the **facts**, is an acyclic logic program [Apt and Bezem, 1991] such that no atomic choice (in an element of $\mathcal{C}_0$) unifies with the head of any rule.

The independent choice logic specifies a particular semantic construction. The semantics is defined in terms of possible worlds. There is a possible world for each selection of one element from each alternative. What follows from these atoms together with $\mathcal{F}$ are true in this possible world.

---
[3] Alternatives correspond to 'variables' in decision theory. This terminology is not used here in order to not confuse logical variables (that are allowed as part of the logic program), and random variables. An atomic choice corresponds to an assignment of a value to a variable; the above definition just treats a variable having a particular value as a proposition (not imposing any particular syntax); the syntactic restrictions and the semantic construction ensure that the values of a variable are mutually exclusive and covering, as well as that the variables are unconditionally independent (see [Poole, 1993])

**Definition 3.3** If $\mathcal{S}$ is a set of sets, a **selector function** on $\mathcal{S}$ is a mapping $\tau : \mathcal{S} \to \cup \mathcal{S}$ such that $\tau(S) \in S$ for all $S \in \mathcal{S}$. The **range** of selector function $\tau$, written $\mathcal{R}(\tau)$ is the set $\{\tau(S) : S \in \mathcal{S}\}$.

**Definition 3.4** Given ICL$_{SC}$ theory $\langle \mathcal{C}_0, A, \mathcal{O}, P_0, \mathcal{F} \rangle$, for each selector function $\tau$ on $\mathcal{C}_0$ there is a **possible world** $w_\tau$. If $f$ is a formula in language $\mathcal{L}_Q$, and $w_\tau$ is a possible world, we write $w_\tau \models f$ (read $f$ is **true in possible world** $w_\tau$) if $\mathcal{F} \cup \mathcal{R}(\tau) \hspace{1pt}\vert\hspace{-4pt}\sim f$.

The existence and uniqueness of the model follows from the acyclicity of the logic program [Apt and Bezem, 1991].

### 3.1 Axiomatising utility

Given the definition of an ICL$_{SC}$ theory, we can write rules for utility. We assume that the utility depends on the situation that the robot ends up in and the possible world. In particular we allow for rules that imply $utility(U, S)$, which is true in a possible world if the utility is $U$ for situation $S$ in that world. The utility depends on what is true in the state defined by the situation and the world — thus we write rules that imply $utility$. This allows for a structured representation for utility. In order to make sure that we can interpret these rules as utilities we need to have utility being functional: for each $S$ there exists a unique $U$ for each world:

**Definition 3.5** An ICL$_{SC}$ theory is **utility complete** if for each possible world $w_\tau$, and each situation $S$ there is a unique number $U$ such that $w_\tau \models utility(U, S)$.

Ensuring utility completeness can be done locally; we have to make sure that the rules for utility cover all of the cases and there are not two rules that imply different utilities whose bodies are compatible.

**Example 3.6** Suppose the utility is the sum of the 'prize' plus the remaining resources:

$$utility(R + P, S) \leftarrow$$
$$prize(P, S) \wedge$$
$$resources(R, S).$$

The prize depends on whether the robot reached its destination or it crashed. No matter what the definition of any other predicates is, the following definition of $prize$ will ensure there is a unique prize for each world and situation[4]:

$$prize(-1000, S) \leftarrow crashed(S).$$
$$prize(1000, S) \leftarrow in\_lab(S) \wedge \sim crashed(S).$$
$$prize(0, S) \leftarrow \sim in\_lab(S) \wedge \sim crashed(S).$$

The resources used depends not only on the final state but on the route taken. To model this we make $resources$ a fluent,

---
[4] We use '$\sim$' to mean negation under Clark completion [Clark, 1978], or in the stable model semantics [Gelfond and Lifschitz, 1988] — these and other semantics for so-called negation as failure coincide for acyclic theories [Apt and Bezem, 1991].

and like any other fluent we axiomatise it:

$$resources(200, s_0).$$
$$resources(R - Cost, do(goto(To, Route), S)) \leftarrow$$
$$\quad at(robot, From, S) \land$$
$$\quad path(From, To, Route, Risky, Cost) \land$$
$$\quad resources(R, S).$$
$$resources(R, do(goto(To, Route), S)) \leftarrow$$
$$\quad crashed(S) \land$$
$$\quad resources(R, S).$$
$$resources(R - 10, do(A, S)) \leftarrow$$
$$\quad \sim gotoaction(A) \land$$
$$\quad resources(R, S).$$
$$gotoaction(goto(A, S)).$$

Here we have assumed that non-goto actions cost 10, and that paths have costs. Paths and their risks and costs are axiomatised using $path(From, To, Route, Risky, Cost)$ that is true if the path from $From$ to $To$ via $Route$ has risk given by $Risky$ can cost $Cost$. An example of this for our domain is:

$$path(r101, r111, direct, yes, 10).$$
$$path(r101, r111, long, no, 100).$$
$$path(r101, r123, direct, yes, 50).$$
$$path(r101, r123, long, no, 90).$$
$$path(r101, door, direct, yes, 50).$$
$$path(r101, door, long, no, 70).$$

### 3.2 Axiomatising Sensors

We also need to axiomatise how sensors work. We assume that sensors are passive; this means that they receive information from the environment, rather than 'doing' anything; there are no sensing 'actions'. This seems to be a better model of actual sensors, such as eyes or ears, and makes modelling simpler than when sensing is an action. So called 'information producing actions' (such as opening the eyes, or performing a biopsy on a patient, or exploding a parcel to see if it is (was) a bomb) are normal actions that are designed to change the world so that the sensors will correlate with the value of interest. Note that under this view, there are no information producing actions, or even informational effects of actions; rather various conditions in the world, some of which are under the robot's control and some of which are not, work together to give varying values for the output of sensors.

Note that a robot cannot condition its action on what is true in the world; it can only condition its actions on what it senses and what it remembers. The only use for sensors is that the output of a sensor depends, perhaps stochastically, on what is true in the world, and thus can be used as evidence for what is true in the world.

Within our situation calculus framework, can write axioms to specify how sensed values depend on what is true in the world. What is sensed depends on the situation and the possible world. We assume that there is a predicate $sense(C, S)$ that is true if $C$ is sensed in situation $S$. Here $C$ is a term in our language, that represents one value for the output of a sensor. $C$ is said to be **observable**.

**Example 3.7** A sensor may be to be able to detect whether the robot is at the same position as the key. It is not reliable; sometimes it says the robot is at the same position as the key when it is not (a false positive), and sometimes it says that the robot is not at the same position when it is (a false negative). The output of the sensor is correlated with what is true in the world, and can be conditioned on in plans.

Suppose that noisy sensor $at\_key$ detects whether the agent is at the same position as the key. For a situation $s$, $sense(at\_key, s)$ is true (in a world) if the robot senses that it is at the key in situation $s$ — the 'at key' sensor returns a positive value— and is false when the robot does not sense it is at the key — the sensor returns a negative value. The $sense(at\_key, S)$ relation can be axiomatised as:

$$sense(at\_key, S) \leftarrow$$
$$\quad at(robot, P, S) \land$$
$$\quad at(key, P, S) \land$$
$$\quad sensor\_true\_pos(S).$$
$$sense(at\_key, S) \leftarrow$$
$$\quad at(robot, P_1, S) \land$$
$$\quad at(key, P_2, S) \land$$
$$\quad P_1 \neq P_2 \land$$
$$\quad sensor\_false\_pos(S).$$

The fluent $sensor\_false\_pos(S)$ is true if the sensor is giving a false-positive value in situation $S$, and $sensor\_true\_pos(S)$ is true if the sensor is not giving a false negative in situation $S$. Each of these could be part of an atomic choice, which would let us model sensors whose errors at different times are independent. The language also lets us write rules for this fluent so that we can model how sensors break.

## 4 GOLOG and Conditional Plans

The idea behind the decision-theoretic planning framework proposed in this paper is that agents get to choose situations (they get to choose what they do, and when they stop), and 'nature' gets to choose worlds (there is a probability distribution over the worlds that specifies the distribution of effects of the actions).

Agents get to choose situations, but they do not have to choose situations blind. We assume that agents can sense the world, and choose their actions conditional on what they observe. Moreover agents can have sequences of acting and observing.

Agents do not directly adopt situations, they adopt 'plans'

or 'programs'. In general these programs can involve atomic actions, conditioning on observations, loops, nondeterministic choice and procedural abstraction. The GOLOG project [Levesque et al., 1996] is investigating such programs. In this paper we only consider simple conditional plans which are programs consisting only of sequential composition and conditioning on observations. One contribution of this paper is to show how conditioning future actions on observations can be cleanly added to GOLOG (in a similar manner to the independently developed robot programs of [Levesque, 1996]).

An example plan is:

$a$; if $c$ then $b$ else $d$; $e$ endIf; $g$

An agent executing this plan will start in situation $s_0$, then do action $a$, then it will sense whether $c$ is true in the resulting situation. If $c$ is true, it will do $b$ then $g$, and if $c$ is false it will do $d$ then $e$ then $g$. Thus this plan either selects the situation $do(g, do(b, do(a, s_0)))$ or the situation $do(g, do(e, do(d, do(a, s_0))))$. It selects the former in all worlds where $sense(c, do(a, s_0))$ is true, and selects the latter in all worlds where $sense(c, do(a, s_0))$ is false. Note that each world is definitive on each fluent for each situation. The expected utility of this plan is the weighted average of the utility for each of the worlds and the situation chosen for that world. The only property we need of $c$ is that its value in situation $do(a, s_0)$ will be able to be observed. The agent does not need to be able to determine its value beforehand.

**Definition 4.1** A **conditional plan**, or just a **plan**, is of the form

$skip$
$A$     where $A$ is a primitive action
$P; Q$  where $P$ and $Q$ are plans
if $C$ then $P$ else $Q$ endIf
        where $C$ is observable; $P$ and $Q$ are plans

Note that '$skip$' is not an action; the $skip$ plan means that the agent does not do anything — time does not pass. This is introduced so that the agent can stop without doing anything (this may be a reasonable plan), and so we do not need an "if $C$ then $P$ endIf" form as well; this would be an abbreviation for "if $C$ then $P$ else $skip$ endIf".

Plans select situations in worlds. We define a relation

$trans(P, W, S_1, S2)$

that is true if doing plan $P$ in world $W$ from situation $S_1$ results in situation $S_2$. This is similar to the $DO$ macro in [Levesque et al., 1996] and the $Rdo$ of [Levesque, 1996], but here what the agent does depends on what it observes, and what the agent observes depends on which world it happens to be in.

We can define the $trans$ relation in pseudo Prolog as:

$trans(skip, W, S, S)$.

$trans(A, W, S, do(A, S)) \leftarrow$
    $primitive(A)$.
$trans((P; Q), W, S_1, S_3) \leftarrow$
    $trans(P, W, S_1, S_2) \land$
    $trans(Q, W, S_2, S_3)$.
$trans((\text{if } C \text{ then } P \text{ else } Q \text{ endIf}), W, S_1, S_2) \leftarrow$
    $W \models sense(C, S_1) \land$
    $trans(P, W, S_1, S_2)$.
$trans((\text{if } C \text{ then } P \text{ else } Q \text{ endIf}), W, S_1, S_2) \leftarrow$
    $W \not\models sense(C, S_1) \land$
    $trans(Q, W, S_1, S_2)$.

Now we are at the stage where we can define the expected utility of a plan. The expected utility of a plan is the weighted average, over the set of possible worlds, of the utility the agent receives in the situation it ends up in for that possible world:

**Definition 4.2** If $\text{ICL}_{SC}$ theory $\langle C_0, A, O, P_0, F \rangle$ is utility complete, the **expected utility** of plan $P$ is[5]:

$$\varepsilon(P) = \sum_\tau p(w_\tau) \times u(w_\tau, P)$$

(summing over all selector functions $\tau$ on $C_0$) where

$u(W, P) = U$ if $W \models utility(U, S)$
    where $trans(P, W, s_0, S)$

(this is well defined as the theory is utility complete), and

$$p(w_\tau) = \prod_{C_0 \in \mathcal{R}(\tau)} P_0(C_0)$$

$u(W, P)$ is the utility of plan P in world $W$. $p(w_\tau)$ is the probability of world $w_\tau$. The probability is the product of the independent choices of nature. It is easy to show that this induces a probability measure (the sum of the probabilities of the worlds is 1).

## 5  Details of our Example

We can model dependent uncertainties. Suppose we are uncertain about whether the door is locked, and where the key is, and suppose that these are not independent, with the following probabilities:

$P(locked(door, s_0)) = 0.9$
$P(at(key, r101, s_0)|locked(door, s_0)) = 0.7$
$P(at(key, r101, s_0)|unlocked(door, s_0)) = 0.2$

(from which we conclude $P(at\_key(r101, s_0)) = 0.65$.)

---
[5]We need a slightly more complicated construction when we have infinitely many worlds. We need to define probability over measurable subsets of the worlds [Poole, 1993], but that would only complicate this presentation.

Following the methodology outlined in [Poole, 1993] this can be modelled as:

$$random([locked(door, s_0) : 0.9,$$
$$unlocked(door, s_0) : 0.1]).$$
$$random([at\_key\_lo(r101, s_0) : 0.7,$$
$$at\_key\_lo(key, r123) : 0.3]).$$
$$random([at\_key\_unlo(r101, s_0) : 0.2,$$
$$at\_key\_unlo(key, r123) : 0.8]).$$
$$at(key, R, s_0) \leftarrow$$
$$at\_key\_lo(R, s_0) \wedge$$
$$locked(door, s_0).$$
$$at(key, R, s_0) \leftarrow$$
$$at\_key\_unlo(R, s_0) \wedge$$
$$unlocked(door, s_0).$$

where $random([a_1 : p_1, \ldots, a_n : p_n])$ means $\{a_1, \ldots, a_n\} \in \mathcal{C}_0$ and $P_0(a_i) = p_i$. This is the syntax used by our implementation.

We can model complex stochastic actions using the same mechanism. The action *goto* is risky; whenever the robot goes past the stairs there is a 10% chance that it will fall down the stairs.

This is modelled with the choice alternatives:

$$random([would\_fall\_down\_stairs(S) : 0.1,$$
$$would\_not\_fall\_down\_stairs(S) : 0.9]).$$

which means

$$\forall S \{would\_fall\_down\_stairs(S),$$
$$would\_not\_fall\_down\_stairs(S)\} \in \mathcal{C}_0$$
$$\forall S\ P_0(would\_fall\_down\_stairs(S)) = 0.1$$

These atomic choices are used in the bodies of rules. We can define the propositional fluent 'at':

$$at(robot, To, do(goto(To, Route), S)) \leftarrow$$
$$at(robot, From, S) \wedge$$
$$path(From, To, Route, no, Cost).$$
$$at(robot, To, do(goto(To, Route), S)) \leftarrow$$
$$at(robot, From, S) \wedge$$
$$path(From, To, Route, yes, Cost) \wedge$$
$$would\_not\_fall\_down\_stairs(S).$$
$$at(robot, Pos, do(A, S)) \leftarrow$$
$$\sim gotoaction(A) \wedge$$
$$at(robot, Pos, S).$$
$$at(X, P, S) \leftarrow$$
$$X \neq robot \wedge$$
$$carrying(robot, X, S) \wedge$$
$$at(robot, P, S).$$
$$at(X, Pos, do(A, S)) \leftarrow$$
$$X \neq robot \wedge$$
$$\sim carrying(robot, X, S) \wedge$$
$$at(X, Pos, S).$$

In those worlds where the path is risky and the agent would fall down the stairs, then it crashes:

$$crashed(do(A, S)) \leftarrow$$
$$crashed(S).$$
$$crashed(do(A, S)) \leftarrow$$
$$risky(A, S) \wedge$$
$$would\_fall\_down\_stairs(S).$$
$$risky(goto(To, Route), S) \leftarrow$$
$$path(From, To, Route, yes, \_) \wedge$$
$$at(robot, From, S).$$

An example plan is:

$$goto(r101, direct);$$
if *at_key*
    then
        $pickup(key);$
        $goto(door, long)$
    else
        $goto(r123, direct);$
        $pickup(key);$
        $goto(door, direct)$
    endIf;
$unlock\_door;$
$enter\_lab$

Given the situation calculus axioms (not all were presented), and the choice space, this plan has an expected utility. This is obtained by deriving $utility(U, S)$ for each world that is selected by the plan, and using a weighted average over the utilities derived. The possible worlds correspond to choices of elements from alternatives. We do not need to generate the possible worlds — only the 'explanations' of the utility [Poole, 1995a]. For example, in all of the worlds where the following is true,

$$\{locked(door, s_0), at(key, r101, s_0),$$
$$would\_not\_fall\_down\_stairs(s_0),$$
$$sensor\_true\_pos(do(goto(r101, direct), s_0)),$$
$$pickup\_succeeds(do(goto(r101, direct), s_0))\}$$

the sensing succeeds (and so the 'then' part of the condition is chosen), the prize is 1000, and the resources left are the initial 200, minus the 10 going from $r111$ to $r101$, minus the 70 going to the door, minus the 30 for the other three actions. Thus the resulting utility if 1090. The sum of the probabilities for all of these worlds is the product of the probabilities of the choices made.

Similarly all of the the possible worlds with $would\_fall\_down\_stairs(s_0)$ true have prize $-1000$,

and resources 190, and thus have utility −810. The probability of all of these worlds sums to 0.1.

The expected utility of this plan can be computed by enumerating the other cases.

## 6  Richer Plan Language

There are two notable deficiencies in our definition of a plan; these were omitted in order to make the presentation simpler.

1. Our programs do not contain loops.

2. There are no local variables; all of the internal state of the robot is encoded in the program counter.

One way to extend the language to include iteration in plans, is by adding a construction such as

while $C$ do $P$ endDo

as a plan (where $C$ is observable and $P$ is a plan), with the corresponding definition of $trans$ being[6]:

$$trans((\text{while } C \text{ do } P \text{ endDo}), W, S_1, S_1) \leftarrow$$
$$W \not\models sense(C, S_1).$$
$$trans((\text{while } C \text{ do } P \text{ endDo}), W, S_1, S_3) \leftarrow$$
$$W \models sense(C, S_1) \land$$
$$trans(P, W, S_1, S_2) \land$$
$$trans((\text{while } C \text{ do } P \text{ endDo}), W, S_2, S_3).$$

This would allow for interesting programs including loops such as

while $everything\_ok$ do $wait$ endDo

(where $wait$ has no effects) which is very silly for deterministic programs, but is perfectly sensible in stochastic domains, where the agent loops until an exogenous event occurs that stops everything being OK. This is not part of the current theory as it violates utility completeness, however, for many domains, the worlds where this program does not halt have measure zero — as long as the probability of failure $> 0$, given enough time something will always break

Local variables can easily be added to the definition of a plan. For example, we can add an assignment statement to assign local variables values, and allow for branching on the values of variables as well as observations. This (and allowing for arithmetic values and operators) will expand the representational power of the language (see [Levesque, 1996] for a discussion of this issue).

---

[6]Note that we really need a second-order definition, as in [Levesque, 1996], to properly define the $trans$ relation rather than the recursive definition here. This will let us characterize loop termination.

The addition of local variables will make some programs simpler, such as those programs where the agent is to condition on previous values for a sensor. For example, suppose the robot's sensor can tell whether a door is unlocked a long time before it is needed. With local variables, whether the door is unlocked can be remembered. Without local variables, that information needs to be encoded in the program counter; this can be done by branching on the sense value when it is sensed, and having different branches depending on whether the door was open or not.

## 7  Comparison with Other Representations

One of the popular action representations for stochastic actions is probabilistic STRIPS [Kushmerick *et al.*, 1995; Draper *et al.*, 1994; Boutilier and Dearden, 1994; Haddawy *et al.*, 1995]. In this section we show that the proposed representation is more concise in the sense that the $\text{ICL}_{SC}$ representation will not be (more than a constant factor) larger than then corresponding probabilistic STRIPS representation plus a rule for each predicate, but that sometimes probabilistic STRIPS representation will be exponentially larger than the corresponding $\text{ICL}_{SC}$ representation.

It is easy to translate probabilistic STRIPS into $\text{ICL}_{SC}$: using the notation of [Kushmerick *et al.*, 1995], each action $a$ is represented as a set $\{\langle t_i, p_i, e_i \rangle\}$. Each tuple can be translated into the rule of form:

$$b_i(a, S) \leftarrow t_i[S] \land r_i[S]$$

($f[S]$ means the state term is added to every atomic formula in formula $f$), where $b_i$ is a unique predicate symbol, the different $r_i$ for the same trigger are collected into an alternative set, such that $P_0(r_i(S)) = p_i$ for all $S$. For those positive elements $p$ of $e_i$, we have a rule:

$$p[do(a, S)] \leftarrow b_i(a, S)$$

For those negative elements $\bar{p}$ of $e_i$ we have the rule,

$$undoes(p, a, S) \leftarrow b_i(a, S)$$

and the frame rule for each predicate:

$$p[do(A, S)] \leftarrow p(S) \land \sim undoes(p, A, S).$$

The $\text{ICL}_{SC}$ action representation is much more modular for some problems than probabilistic STRIPS. As in STRIPS, the actions have to be represented all at once. Probabilistic STRIPS is worse than the $\text{ICL}_{SC}$ representation when actions effect fluents independently. At one extreme (where the effect does not depend on the action), consider stochastic 'frame axioms' such as the axiom for *carrying* presented in Example 2.3. In probabilistic STRIPS the conditional effects have to be added to every tuple representing an action — in terms of [Kushmerick *et al.*, 1995], for every trigger that is compatible with carrying the key, we have to split into the cases where the agent drops the key and the agent doesn't. Thus the probabilistic STRIPS representation grows exponentially with the number of independent stochastic frame axioms: consider $n$ fluents which persist stochastically and independently and the *wait* action, with no effects. The $ICL_{SC}$ representation is linear in the number of fluents, whereas the probabilistic STRIPS representation is exponential in $n$. Note that if the persistence of the fluents are not independent, then the $ICL_{SC}$ representation will also be the exponential in $n$ — we cannot get better than this; the number of probabilities that have to be specified is also exponential in $n$. In some sense we are exploiting the conciseness of Bayesian networks — together with structured probability tables (see [Poole, 1993]) — to specify the dependencies amongst the outcomes.

The $ICL_{SC}$ representation is closely related to two slice temporal Bayesian networks [Dean and Kanazawa, 1989] or the action networks of [Boutilier et al., 1995; Boutilier and Poole, 1996] that are used for Markov decision processes (MDPs). The latter represent in trees what is represented here in rules — see [Poole, 1993] for a comparison between the rule language presented here and Bayesian networks. The situation calculus rules can be seen as structured representations of the state transition function, and the rules for utility can be seen as a structured representation of the reward or value function. In [Boutilier and Poole, 1996], this structure is exploited for finding optimal policies in partially observable MDPs. A problem with the POMDP conception is that it assumes that agents maintain a belief state (a probability distribution over possible worlds). In order to avoid this, POMDP researchers (see [Kaelbling et al., 1996]) have proposed 'policy trees', which correspond to the plans developed here. The general idea behind the structured POMDP algorithm [Boutilier and Poole, 1996] is to use what is essentially regression [Waldinger, 1977] on the situation calculus rules to build plans of future actions contingent on observations — policy trees. The difficult part for exact computation is to not build plans that are stochastically dominated [Kaelbling et al., 1996]. One problem with the action networks is that the problem representations grow with the product of the number of actions and the number of state variables — this is exactly the frame problem [McCarthy and Hayes, 1969] that is 'solved' here using Reiter's solution [Reiter, 1991]; if the number of actions that affect a fluent is bounded, the size of the representation is proportional the number of fluents (state variables).

In contrast to [Haddawy and Hanks, 1993], we allow a general language to specify utility. Utility can be an arbitrary function of the final state, and because any information about the past can be incorporated into the state, we allow the utility to be an arbitrary function of the history. The aim of this work is not to identify useful utility functions, but rather to give a language to specify utilities.

The use of probability in this paper should be contrasted to that in [Bacchus et al., 1995]. The agents in the framework presented here do not (have to) do probabilistic reasoning. As, for example in MDPs, the probabilistic reasoning is about the agent and the environment. An optimal agent (or an optimal program for an agent) may maintain a belief state that is updated by Bayes rule or some other mechanism, but it does not have to. It only has to do the right thing. Moreover we let the agent condition its actions based on its observations, and not just update its belief state. We can also incorporate non-deterministic actions.

## 8 Conclusion

This paper has presented a formalism that lets us combine situation calculus axioms, conditional plans and Bayesian decision theory in a coherent framework. It is closely related to structured representations of POMDP problems. The hope is that we can form a bridge between work in AI planning and in POMDPs, and use the best features of both. This is the basis for ongoing research.

The way we have treated the situation calculus (and we have tried hard to keep it as close to the original as possible) really gives an agent-oriented view of time — the 'situations' in some sense mark particular time points that correspond to the agent completing its actions. Everything else (e.g., actions by nature or other agents) then has to meld in with this division of time. Whether this is preferable to a more uniform treatment of the agent's actions and other actions (see e.g., [Poole, 1995b]) is still a matter of argument.